\documentclass[sigconf,natbib=false,nonacm]{acmart} 
\AtBeginDocument{%
  }

\usepackage{bm}
\usepackage{booktabs}
\usepackage{xcolor}
\usepackage{amsmath}
\usepackage{bm}
\usepackage{amsfonts}
\usepackage{multicol}
\usepackage{graphicx}
\usepackage{multirow}
\usepackage{graphicx} 
\usepackage{wrapfig}
\usepackage{caption}
\usepackage{subcaption}
\usepackage{adjustbox}
\usepackage{enumitem}
\usepackage{booktabs}
\usepackage{array}
\usepackage{pifont}
\usepackage{balance}

\usepackage{hyperref}

\usepackage{listings}

\usepackage{graphicx}
\usepackage{subcaption}

\RequirePackage[
  datamodel=acmdatamodel,
  style=acmnumeric,
  sorting=none,
]{biblatex}

\begin{document}

\title{Risk In Context: Benchmarking Privacy Leakage of Foundation Models in Synthetic Tabular Data Generation}


\author{Jessup Byun, Xiaofeng Lin, Joshua Ward, Guang Cheng}
\email{{jessupbyun, bernardo1998, joshuaward, guangcheng}@g.ucla.edu}
\affiliation{%
    \institution{University of California, Los Angeles}
    \city{Los, Angeles}
    \state{CA}
    \country{USA}
}


\begin{abstract}
Synthetic tabular data is essential for machine learning workflows, particularly as they expand small or imbalanced datasets and facilitate privacy-preserving data sharing. Yet, state-of-the-art generative models (GANs, VAEs, and diffusion models) achieve their promised fidelity only with large datasets containing thousands of examples. In low-data scenarios—often the primary motivation for using synthetic data—these models frequently overfit, leak sensitive records, and require constant retraining as new data arrive. To address these limitations, recent studies have leveraged large pre-trained transformers capable of generating new data rows via \emph{in-context learning (ICL)}. These foundation models require only a few seed examples to produce additional records without any parameter updates, thus overcoming both data scarcity and retraining bottlenecks. However, in-context learning (ICL) repeats the seed rows verbatim at each generation step, creating a novel privacy threat that has so far been quantified only for text.  How serious this leakage is for tabular synthesis—where a single row can uniquely identify an individual—remains unknown.  

We close this gap with the first end-to-end benchmark of three foundation-model generators—GPT-4o-mini, LLaMA 3.3 70B, and TabPFN v2—against four state-of-the-art baselines on 35 real-world tables spanning health, finance, and public policy. Evaluated on statistical fidelity, downstream utility, and worst-case membership inference leakage, the study reveals that foundation models consistently occupy the upper end of the privacy-risk spectrum. We find that LLaMA 3.3 70B poses the highest privacy risk, yielding up to 54 percentage points (pp) higher true-positive rate (at 1\% FPR) than the safest deep-learning baseline. GPT-4o-mini and TabPFN also rank among the most vulnerable models, revealing the elevated leakage potential of foundation models. Quality–leakage tradeoff plots illustrate a privacy–utility frontier, with CTGAN and GPT-4o-mini offering favorable balances. A factorial study demonstrates that three zero-cost prompt-level mitigations—small batch size, low (but non-zero) temperature, and inclusion of summary statistics—can reduce worst-case AUC by 14 pp and rare-class leakage by up to 39 pp, while retaining over 90\% of baseline fidelity. Our benchmark and mitigation recipe provide an actionable blueprint for safer, low-data tabular synthesis using foundation models.

\end{abstract}

\maketitle


\section{Introduction}

Tabular data underpin a vast share of real-world machine-learning pipelines, from clinical registries to credit-risk engines.  
Consequently, \emph{synthetic tabular data}—records drawn from a model that approximates the joint distribution of the original table—has become a key instrument for modern ML workflows~\cite{fonseca2023tabularSynthetic}.  
Its value is two-fold.  
First, synthetic rows can augment \emph{small or imbalanced datasets}, bolstering the statistical power of models trained on rare-disease cohorts or under-represented demographic groups.  
Second, because the generated rows do not correspond to real individuals, they enable data sharing while preserving privacy, a requirement that is particularly stringent in healthcare~\cite{vallevik2024canHealthcare,hernandez2022syntheticHealth} and finance~\cite{wu2023interpretationVAEfinance}.  
By offering both stronger sample support and reduced disclosure risk~\cite{platzer2021holdout,mckenna2022aim}, synthetic tables now play a fundamental role in scaling machine learning to domains where real data are scarce or legally restricted.


However, deep table generators themselves are data-hungry learning algorithms.  
Empirical studies show that state-of-the-art GAN, VAE, and diffusion-based models for tabular data~\cite{xu2019modeling,kim2021oct,zhao2024ctabplus,kotelnikov2023tabddpm,zhang2024mixedtype} attain their advertised fidelity only when thousands of training rows are available.  
In the very regimes where synthetic data are most demanded—rare-disease cohorts or minority-group slices with \(\,<\!500\) real examples—these generators (i) overfit, reproducing near-duplicates that endanger privacy\,\cite{vanbreugel2023membership}, and (ii) deliver little utility because the synthetic set lacks diversity\,\cite{cannon2025structured}.  
Moreover, as production streams may append millions of new rows every hour, retraining such \emph{data-specific} models becomes prohibitively slow\,\cite{kotelnikov2023tabddpm}, undermining their scalability in real deployments.

To overcome these limitations, researchers have turned to \emph{large pre-trained transformers}—hereafter large language models (LLMs) and \emph{foundation models for tabular data}.  
Such models, pre-trained on massive corpora, can generate new rows \emph{in context}: given only a handful of seed examples, they sample additional records without any weight updates.  
This in-context learning (ICL) regularizes generation in low-data regimes, yielding realistic and diverse rows while eliminating the retraining bottleneck.  
Two representative families are (i) \emph{general-purpose language models} prompted with column headers and a few exemplary rows~\cite{li2023syntheticLLM,seedatcurated}, and (ii) \emph{tabular-specific transformers} pre-trained across thousands of small tables, such as TabPFN~\cite{hollmann2022tabpfn, hollmann2025accurate}.

Despite its promise, \emph{in-context learning (ICL)} introduces significant privacy concerns. Specifically, the small number of real data rows embedded in the prompt are repeatedly exposed verbatim to the model during each generation step. Since the model undergoes no parameter updates, it is free to reproduce these seed rows token-by-token—a vulnerability recently documented in the context of large language models (LLMs). Wen \textit{et al.} demonstrated that membership-inference attacks can achieve over 95\% accuracy against several open-source LLMs, relying solely on observing generated outputs~\cite{wen2024iclmia}. Additionally, Nakka \textit{et al.} showed that multi-query attacks could increase the extraction rate of personally identifiable information by up to five times~\cite{nakka2025piiscope}. Even proposed methods that incorporate differential privacy directly into the prompts have confirmed that naive ICL approaches remain susceptible to substantial row-level privacy leakage~\cite{tang2024dpicl}.

In the case of tabular data, the privacy risks are further exacerbated. Each record often represents a unique combination of quasi-identifiers—attributes such as rare disease diagnoses, ZIP codes, age, or income—which makes any verbatim reproduction immediately identifiable. However, a comprehensive and systematic evaluation quantifying the precise privacy risks associated with ICL specifically for synthetic tabular data generation has yet to be conducted.

\vspace{0.3em}

\noindent{Our contributions are as follows:}
\begin{itemize}[leftmargin=1.2em,itemsep=0.4em]
    \item \textbf{Holistic benchmark of privacy in low-data table synthesis.}  
          We compare three foundation-model ICL generators (GPT-4o-mini, LLaMA 3.3 70B, and TabPFN v2) with four state-of-the-art
          data-specific baselines (CTGAN, TVAE, TabDiff, and SMOTE) on \emph{35} real-world tables
          spanning health, finance, and public-policy domains.
          Evaluated on statistical fidelity, downstream utility, and worst-case membership-inference leakage,
          the study reveals that foundation models occupy the upper end of the privacy-risk spectrum:
          Llama 3.3 70B yields up to \textbf{54\%} higher true-positive rate (at 1\% FPR) than the safest deep-learning baseline,
          while GPT-4o-mini and TabPFN sit near the high end of the deep-generator cluster.
          These cross-model, cross-dataset leakage patterns have not been quantified before.

\item \textbf{Prompt-level privacy knobs and trade-off frontier.}  
      A factorial study of three zero-cost prompt parameters—(i) generation
      batch size, (ii) sampling temperature, and (iii) inclusion of summary
      statistics—maps how simple edits shift foundation models along the
      privacy–utility frontier.  Tuning these levers cuts worst-case AUC by
      up to 14 pp and slashes rare-class leakage by 24–39 pp, while
      preserving >90\% of baseline fidelity. The recipe of \emph{small batch,
      low (but non-zero) temperature, and explicit summary statistics} is an
      immediately deployable defense that requires no retraining.
\end{itemize}

\section{Related Work}

\subsection{Synthetic Tabular Data Generation}

Early work on STDG was rooted in classical statistical techniques such as CART synthesis, parametric bootstrap, and SMOTE-based resampling~\cite{reiter2005usingCART,chawla2002smote,nowok2016synthpop}.  
While effective for small, low-dimensional tables, these approaches struggle to capture complex multivariate dependencies.

\textbf{Deep generative models.}  
Recent deep‐learning methods have markedly expanded the scope of STDG~\cite{che2017boosting,figueira2022surveysdg}.  
CTGAN and TVAE~\cite{xu2019modeling} marry conditional generation with Generative Adversarial Networks (GANs) and Variational Autoencoders (VAEs) to cope with imbalanced, heavy‐tailed columns.  
CtabGAN+~\cite{zhao2021ctab,zhao2024ctabplus} extends this idea to mixed‐type variables with long‐tailed distributions, whereas \emph{TabDDPM}~\cite{kotelnikov2023tabddpm} achieves state‐of‐the‐art fidelity by running separate diffusion processes for numerical versus categorical features.  
\emph{Autodiff}~\cite{suh2023autodiff} and \emph{Tabsyn}~\cite{zhang2024mixedtype} follow a similar diffusion paradigm but rely on table‐specific autoencoders, limiting their cross‐domain transferability.

\textbf{Privacy‐aware generators.}  
A parallel line of work incorporates Differential Privacy (DP) into the training objective~\cite{jordon2018pate,zhang2017privbayes,mckenna2022aim}.  
Although DP provides formal guarantees, existing DP generators often sacrifice either statistical fidelity or downstream utility, failing to improve model performance when the synthetic data are used for augmentation~\cite{manousakas2023usefulness}.

\textbf{Foundation‐model approaches.}  
Prompt‐fine-tuning methods such as \emph{GReaT}~\cite{borisov2022language} and \emph{Tabula}~\cite{zhao2023tabula} first linearize each row into natural-language text and then \emph{continue training} the language model.  
Although effective, they require gradient updates and large GPU budgets.

A parallel line of work keeps the backbone \emph{frozen} and relies purely on \emph{in-context learning}. Curated LLM~\cite{seedatcurated} generates synthetic tables in low-data regimes combining prompting LLM API with example rows and column statistics, and curating feature relationships in synthesized tables via learning dynamics of oracle models. Kim \textit{et al.} introduces EPIC~\cite{kim2025epic}, a grouped-CSV prompting scheme that lets an off-the-shelf LLM synthesize balanced data for minority classes without any fine-tuning. For tabular-specific transformers, TabPFN v2~\cite{hollmann2025accurate} performs tabular synthesis by sequential in-context prediction of the next feature given generated ones. Ma \textit{et al.} turns the pre-trained TabPFN into an energy-based generator dubbed TabPFGen~\cite{ma2024tabpfgen}; sampling is performed with Langevin dynamics, again requiring no additional training.  
These ICL-only techniques demonstrate that foundation models can act as drop-in tabular generators, yet their privacy characteristics remain largely unexplored.

Overall, while modern deep generative models significantly outperform classical statistical methods, they remain impractical due to the constant retraining required for each new dataset. Meanwhile, foundation-model approaches present new challenges, particularly numerical precision issues and increased privacy risks—these are precisely the gaps our work addresses.

\subsection{Privacy Auditing of Tabular Synthetic Data With Membership-Inference Attack}

 Membership Inference Attacks (MIAs) aim to classify whether a specific observation was a member of the original dataset used to train a model. Let \(X\) be a random variable on domain \(\mathcal{X}\) with distribution \(p_X(X)\), and $T$ be a dataset of independent samples from \(p_X(X)\). A generative model \(G\), trained on $T$, then generates a synthetic dataset $S$. An adversary \(\mathcal{A}: X \to \{0, 1\}\) aims to determine if a test sample \(x^*\) is an element of $T$: $T \sim p_X(X)$. Formally, this classification or Membership Inference Attack can be expressed as:
\begin{equation}\label{eq:membership_prediction}
    \mathcal{A}(x^{\star}) = \mathbb{I}\left[f(x^{\star}) > \gamma\right]
\end{equation}
where $\mathbb{I}$ is the indicator function, $f(x^{\star})$ is a scoring function of the test observation $x^*$, and $\gamma$ is an adjustable decision threshold. The success of the attack can be measured using traditional binary classification metrics and can be interpreted as a measure of the privacy leakage from a model of the training data.

 To construct their attack, the adversary relies on some prior given information called a threat model. These include black box attacks \cite{Hayes2017LOGANMI, Hilprecht2019MonteCA, ganleaks} in which only $S$ is available,
  shadow box (also called calibrated) attacks in which both $S$ and then a reference dataset $R$ from the same population distribution of the training set are given \cite{vanbreugel2023membership,ward2024dataplagiarismindexcharacterizing}, and white box attacks \cite{sablayrolles2019white} in which both $S$, $R$ and full access to the model are known. Other lines of work have explored threat models where the adversary assumes a shadow-box threat model but additionally knows the implementation, but not the training weights, of the tabular generator \cite{groundhog, houssiau2022tapas,Meeus_2024}.

 MIAs leverage information from a specified threat model along with some observation regarding model failure behavior to exploit potential vulnerabilities in constructing Equation \ref{eq:membership_prediction}. 
 For example, a variety of attacks from \cite{ganleaks} and  \cite{houssiau2022tapas} target memorization by computing the distance between $x^*$ and the closest observation from $S$. Other MIAs focus on overfitting, where the model produces synthetic samples that are too similar in distribution to the training dataset relative to the overall population distribution. Methods such as DOMIAS \cite{vanbreugel2023membership} and DPI \cite{ward2024dataplagiarismindexcharacterizing} attack overfitting by comparing the density of synthetic observations in a local region to that of a reference dataset. 

While methodologically diverse, MIAs targeting synthetic data aim to uncover the same fundamental issue: the potential for generative models to inadvertently reveal information about their training data. If a model produces synthetic records that allow an adversary to infer training membership, it constitutes a direct breach of privacy. This leakage signals a failure in the model, as it indicates an imbalance between generating realistic data and preserving confidentiality. A well-calibrated generative model should neither reproduce training samples nor generate synthetic data that is overly concentrated around specific regions of the training distribution.

\section{Methodology}
\label{sec:method}

Our benchmark evaluates a spectrum of tabular–data generators under a unified
low–data protocol.  Below we detail (i) the models under comparison,
(ii) the dataset and sampling procedure, and (iii) the evaluation metrics
used to quantify fidelity, downstream utility, and privacy leakage.

\subsection{Synthetic–Data Generators}
\label{sec:generators}

\textbf{(A) Foundation models \emph{via} in-context learning (ICL).}
\begin{itemize}[leftmargin=1.1em,itemsep=0.1em]
    \item \textbf{TabPFN v2}~\cite{hollmann2025accurate}: a transformer pre-trained across millions of synthetic tables for few-shot tabular prediction. We adapt it to generation by autoregressively sampling each feature conditioned on previously sampled features.
    \item \textbf{GPT-4o mini}~\cite{openai2024gpt4omini}:
          a general-purpose LLM queried using a structured prompt that specifies the dataset name, schema (column names and ordering), statistical summaries (numerical and categorical), and a sample of the real data. The prompt includes a complete CSV-formatted snippet of the training data and instructs the model to generate new samples with matching structure and diversity. To encourage faithful yet varied outputs, we also describe the generation task as mimicking causal and statistical properties while explicitly avoiding extra columns or formatting artifacts. The full prompt is detailed in Appendix~\ref{sec:prompt}.
    \item \textbf{LLaMA 3.3 70B}~\cite{meta2024llama3_3_70b}: an open-source LLM prompted analogously to GPT-4o mini, enabling head-to-head comparison between closed-and open-source LLMs.
\end{itemize}

\noindent{\textbf{(B) Data-specific generators (trained per dataset).}}
\begin{itemize}[leftmargin=1.1em,itemsep=0.1em]
    \item \textbf{TabDiff}~\cite{shi2024tabdiff}: a joint continuous‐time diffusion model for mixed‐type tabular data that defines feature‐wise learnable diffusion processes to capture the heterogeneity of numerical and categorical columns.
    \item \textbf{CTGAN} and \textbf{TVAE}~\cite{xu2019modeling}: CTGAN employs a conditional GAN with mode‐specific normalization layers using a Gaussian mixture model to encode continuous features and conditional vectors for categorical features, plus “training‐by‐sampling” to improve mode coverage; TVAE uses a variational autoencoder to embed mixed‐type features into a continuous latent space and decodes via mode‐specific normalization, optimizing the ELBO with structural regularization to balance fidelity and diversity.
    \item \textbf{SMOTE}~\cite{chawla2002smote}: a k‐nearest‐neighbor oversampling technique that generates synthetic minority‐class samples by interpolating between each sample and its randomly selected nearest neighbors in feature space.
\end{itemize}

\subsection{Datasets and Low-Data Protocol}
\label{sec:data}

\textbf{Public benchmark.}
We use the OpenML \textsc{ctr23} suite~\cite{fischer2023openmlctr}, comprising of
35 classification tables with
heterogeneous schema. The suite comprises 35 real-world tables spanning 500–100,000 rows and no more than 5000 engineered features after one-hot encoding; numerical attributes dominate, but every dataset also includes several categorical columns, giving the heterogeneous structure that we target. 

\textbf{Splitting and subsampling.}
For each dataset, we create an
$80\!:\!20$ train–test split on the real data, stratified based on target classes.
To emulate low-data regimes, we draw
$\{32,\,64,\,128\}$ training rows without replacement and
repeat the draw with random seeds $0,1,2$.
ICL models receive the sampled rows as exemplars;
trainable generators fit their parameters on the same subsample.

\subsection{Evaluation Metrics}
\label{sec:metrics}

We benchmark every generator along three complementary axes:
(\emph{i})~statistical fidelity of the synthetic table,
(\emph{ii})~downstream utility in real predictive tasks, and
(\emph{iii})~privacy leakage against membership-inference attacks (MIAs).

\subsubsection{Statistical Fidelity}
\label{sec:fidelity}
For statistical fidelity, we measure both the marginal column distribution as well as joint distribution. 
For numerical columns, we compute the Kolmogorov–Smirnov (KS) distance between
real and synthetic marginals; for categorical columns, we use the
$\chi^2$ divergence on contingency tables. We subtract the column distance from 1 so that larger values indicate smaller distance and thus better marginal distribution modeling. 
For joint distribution, we use similarity of column correlation in real and synthetic data. We calculate Pearson's correlation for numerical-numerical column pairs; for categorical-categorical pairs, we compute a normalized contingency table for the real and synthetic data. This table describes the proportion of rows that have each combination of categories in A and B.
Then, it computes the difference between the contingency tables using the Total Variation Distance. Finally, we subtract the distance from 1 to ensure that a high score means high similarity. 
For numerical-categorical pairs, we convert the numerical column into bins by quantile and compute the contingency similarity. We followed the implementation of SDMetrics~\cite{sdmetrics} of fidelity metrics.

\subsubsection{Downstream Utility}
\label{sec:utility}
To evaluate the machine learning utility of synthetic data, we fit the following classifiers: logistic regression, Naïve Bayes, decision tree, random forest, XGBoost~\cite{chen2016xgboost}, and CatBoost~\cite{prokhorenkova2018catboost}, and then evaluate them on the holdout test set with a macro-average ROC AUC score. For 

\subsubsection{Privacy Leakage}
\label{sec:privacy}

\paragraph{Threat model.}
We consider black-box and model-unknown shadow-box
adversaries who see the released synthetic set \(S\)—
optionally supplemented by a reference set \(R\)—
but never the generator code or parameters.
This mirrors realistic data-sharing scenarios and enables model-agnostic,
computationally feasible audits~\cite{vanbreugel2023membership,golob2024privacyvulnerabilitiesmarginalsbasedsynthetic}.

\paragraph{Auditing procedure.}
Following the empirical-worst-case principle of
Empirical Differential Privacy (EDP)~\cite{Jagielski2020},
we measure leakage as the \emph{maximum} attack AUC across
\(\mathcal{A}=13\) state-of-the-art MIAs that span distance, density,
and classifier signals:
\[
\text{Leakage} \;=\;
\max_{A\in\mathcal{A}} \text{AUC}(A).
\]
The Area Under Receiver Operating Characteristic curve (ROC AUC) and True Positive Rate (TPR) at False Positive Rate (FPR) are used as performance metrics of attacks, where a higher performance indicates more success in identifying membership of training data points, and thus greater privacy leakage. All attacks are re-implemented in a common Python framework; no model
re-training is required.  Table~\ref{tab:mia_list} lists the methods. We used the Synth-MIA~\cite{ward2025synthmia} library as our attack framework implementation. 

\begin{table}[ht]
\centering
\caption{Membership-inference attacks used in this study.}
\label{tab:mia_list}
\begin{tabular}{lll}
\toprule
\textbf{Attack} & \textbf{Threat model} & \textbf{Signal type} \\
\midrule
DOMIAS \cite{vanbreugel2023membership}           & Shadow-box & Density ratio \\
DPI \cite{ward2024dataplagiarismindexcharacterizing}              & Shadow-box & Local density \\
Classifier \cite{houssiau2022tapas}      & Shadow-box & Density ratio \\
Density Estimator \cite{houssiau2022tapas} & Black-box & Density estimation \\
DCR \cite{ganleaks}                     & Black-box & Distance-based \\
DCR-Diff \cite{ganleaks}                & Shadow-box & Distance difference \\
Logan \cite{Hayes2017LOGANMI}            & Shadow-box & Density ratio \\
MC Estimation \cite{Hilprecht2019MonteCA} & Black-box & Density estimation \\
\bottomrule
\end{tabular}

\end{table}

\section{Experiments}

\subsection{Privacy Leakage}
\label{sec:privacy_results}

Table~\ref{tab:privacy_scores} reports the \emph{worst-case} membership-inference
performance, averaged over all datasets, subset sizes, and random seeds—for each
generator class. Two consistent patterns emerge:

\vspace{0.3em}
\noindent\textbf{Foundation models sit at the higher end of the privacy-risk spectrum.}
The three foundation-model variants occupy the upper end of the leakage range:
their mean worst-case AUCs fall in the range
$0.587$–$0.667$, compared with $0.580$–$0.627$ for CTGAN, TVAE, and TabDiff.
At an operational false-positive rate of $1\%$, the corresponding
$\text{TPR}_{0.01}$ climbs from \(\,0.042\)–\(0.061\) (deep generators)
to \(\,0.054\)–\(0.181\) (foundation models), indicating that an adversary can
recover up to three times as many true members with high confidence when attacking a foundation-model
synthesizer.  Nonetheless, all deep-learning generators remain substantially
safer than the interpolation-based baseline SMOTE, whose AUC of
$0.831$ and $\text{TPR}_{0.01}=0.438$ confirm that naïve interpolation-based up-sampling of minority
rows is highly vulnerable.

\vspace{0.3em}
\noindent\textbf{Large language models vary widely in leakage.}
Within the foundation-model family, leakage is \emph{not uniform}.
LLaMA 3.3 70B shows the highest risk (AUC $0.667$,
$\text{TPR}_{0.1}=0.345$), exceeding even the safest data-specific generator
by roughly \(+0.09\) pp.  By contrast,
TabPFN v2 ($0.620$) and GPT-4o mini ($0.587$) hover at the
upper end of the deep-generator cluster, suggesting that model architecture,
prompt format, or pre-training corpus strongly influences membership exposure.
This disparity underscores the need for per-model audits rather than blanket
assumptions about “foundation-model risk.”

\begin{table*}[tb]
  \centering
  \small
  \caption{Privacy leakage metrics: mean (standard deviation) of the worst‐case attacker performance across all dataset, subset sizes and seeds.}
  \renewcommand{\arraystretch}{1.0} 
  \resizebox{\textwidth}{!}{%
    \begin{tabular}{lccccc}
      \toprule
      Model        & AUC            & TPR@FPR=0      & TPR@FPR=0.001   & TPR@FPR=0.01    & TPR@FPR=0.1     \\
      \midrule
      SMOTE        & 0.831 (0.064)  & 0.273 (0.211)  & 0.310 (0.202)   & 0.438 (0.181)   & 0.648 (0.136)   \\
      \hline
      CTGAN        & 0.580 (0.047)  & 0.008 (0.016)  & 0.012 (0.018)   & 0.042 (0.028)   & 0.185 (0.058)   \\
      TVAE         & 0.627 (0.065)  & 0.017 (0.033)  & 0.023 (0.034)   & 0.061 (0.048)   & 0.244 (0.087)   \\
      TabDiff      & 0.600 (0.062)  & 0.012 (0.024)  & 0.016 (0.024)   & 0.052 (0.038)   & 0.213 (0.083)   \\
      \hline
      GPT-4o-mini  & 0.587 (0.052)  & 0.016 (0.027)  & 0.023 (0.030)   & 0.054 (0.038)   & 0.205 (0.074)   \\
      TabPFN v2      & 0.620 (0.059)  & 0.015 (0.025)  & 0.023 (0.028)   & 0.060 (0.038)   & 0.243 (0.078)   \\
      LLaMA 3.3 70B   & \textbf{0.667} (0.128)  & \textbf{0.100} (0.224)  & \textbf{0.117} (0.236)   & \textbf{0.181} (0.264)   & \textbf{0.345} (0.240)   \\
      \bottomrule
    \end{tabular}%
  }
  \label{tab:privacy_scores}
\end{table*}

\subsection{Impact of Dataset Size}
\label{sec:privacy_size}

Figure~\ref{fig:tradeoff_all} illustrates how the mean worst-case AUC varies
when the number of real rows supplied to each generator is increased from
\(n=32\) to \(n=128\).
Three key observations stand out:

\begin{enumerate}[leftmargin=1.2em,itemsep=0.2em]
\item \textbf{Smaller samples leak more.}
      Every \emph{learned} generator—GAN, VAE, diffusion, and foundation
      model—shows a monotone decrease in leakage as the training subset grows.
      For instance, TVAE drops from \(0.668\) (\,\(n=32\)) to \(0.587\)
      (\,\(n=128\)), while GPT-4o mini falls from \(0.617\) to \(0.563\).
      The trend supports the intuition that with fewer examples the model must
      rely on memorisation, thereby exposing members more readily.

\item \textbf{Foundation-model dispersion persists across sizes.}
      Although leakage lessens with data, the ranking among foundation models
      remains: LLaMA 3.3 70B is consistently the riskiest
      (AUC \(0.713 \rightarrow 0.625\)),
      whereas TabPFN v2 and GPT-4o mini track the upper
      bound of the deep-generator cluster.

\item \textbf{SMOTE is size-insensitive yet unsafe.}
      The interpolation baseline stays near \(0.83\) AUC regardless of \(n\),
      confirming that simply copying or perturbing minority rows offers no
      privacy benefit even when more seeds are available.
\end{enumerate}

\noindent
\textbf{Implication.}
Low-data scenarios—precisely the cases where synthetic augmentation is most
needed—also pose the greatest privacy exposure, particularly for large LLMs.
Mitigation efforts should therefore prioritize these extreme regimes.

\begin{figure*}[!htbp]
  \centering
  \includegraphics[width=\linewidth]{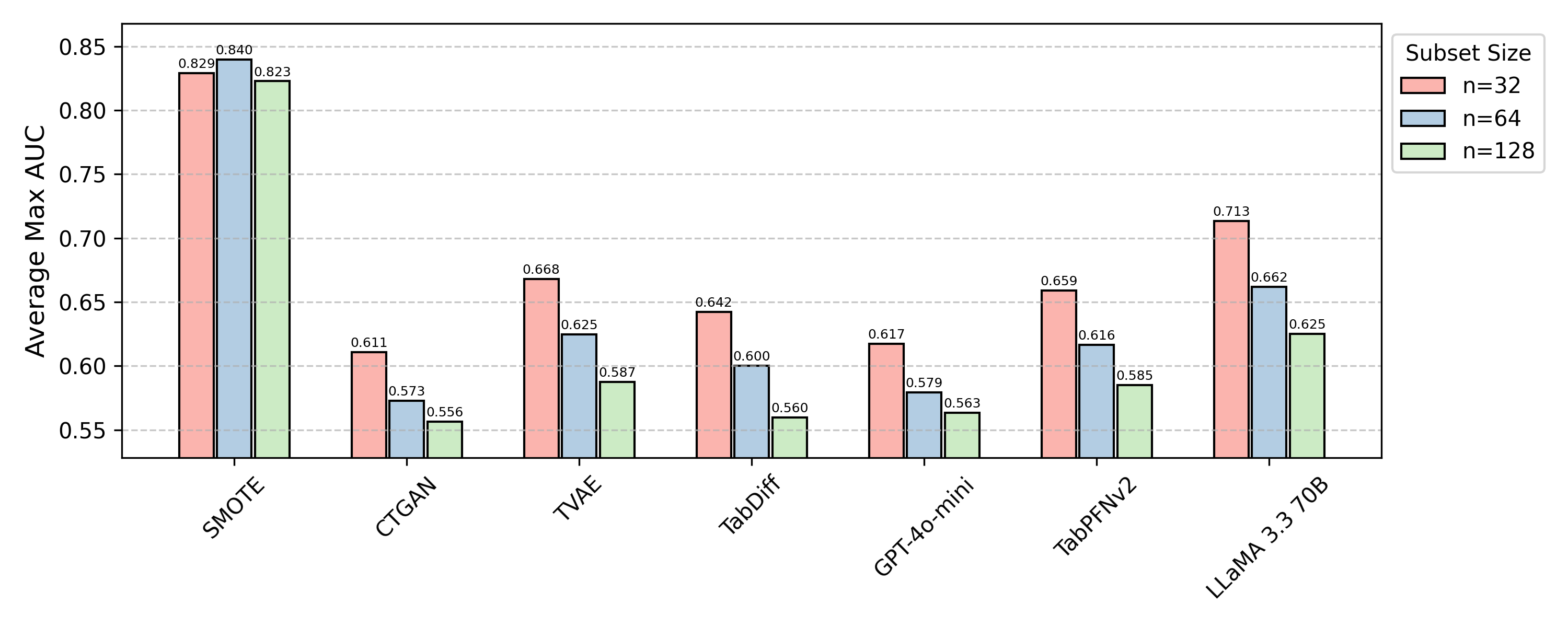}
  \caption{Privacy leakage (mean worst‐case AUC) for each synthetic data generator across three subset sizes.  
  For each model we compute the maximum AUC over all attackers for each of the 315 splits (35 datasets $\times$ 3 seeds $\times$ 3 sizes), then show the mean of those 315 worst‐case values. Bars are color‐coded by subset size ($n=32,64,128$).}
  \label{fig:privacy_leakage_sizes}
\end{figure*}

\subsection{Trade-off between Privacy and Data Quality}

\begin{figure*}[!htbp]
  \centering
  \includegraphics[width=0.9\textwidth]{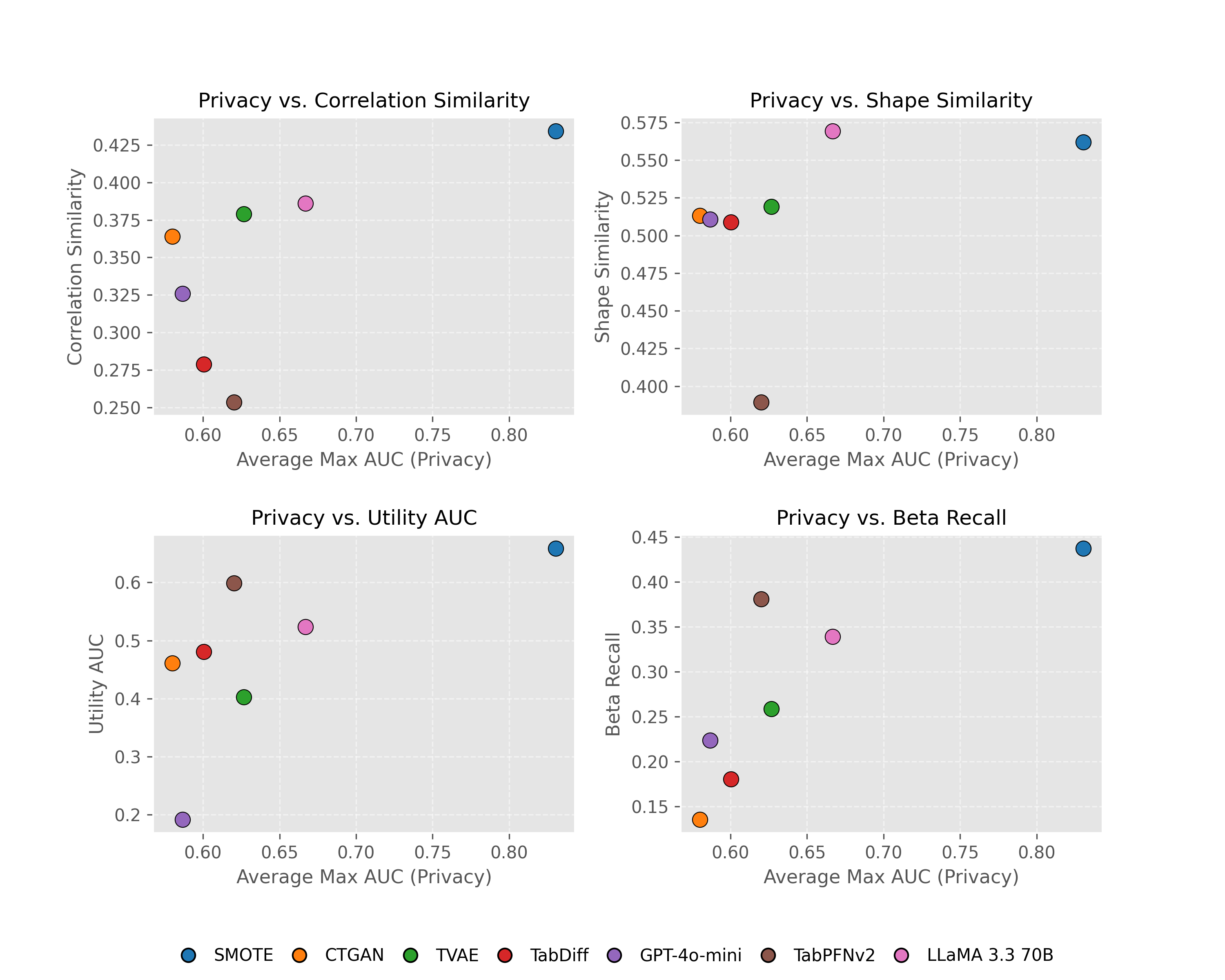}
  \caption{%
    Privacy–utility trade‐off across synthetic generators.  
    Each point represents one generator’s \emph{mean} worst‐case membership‐inference AUC
    (x‐axis: Average Max AUC, Privacy) plotted against its \emph{mean}
    fidelity/diversity/utility score (y‐axis: metrics), for correlation similarity,
    shape similarity, utility AUC, and beta recall in a 2×2 grid.
    Points are colored by model; legend at bottom. Error bars omitted for clarity.
  }
  \label{fig:tradeoff_all}
\end{figure*}

\paragraph{Privacy--Utility Trade-off.}
Figure~\ref{fig:tradeoff_all} juxtaposes the mean worst-case privacy AUC
($x$-axis) with four quality indicators ($y$-axis):
(column-pair) \textit{correlation similarity},
\textit{column shape similarity}, downstream \textit{classifier AUC},
and \textit{beta recall}~\cite{alaa2022faithful} (diversity).
Across all panels we observe a clear, approximately monotone frontier:
models that score higher on any quality metric
also suffer larger membership-inference leakage.
Put differently, \emph{better data quality lines up with weaker privacy}. 

\noindent
\textbf{Deviation from the naive $y=x$ frontier.}
In an ideal scenario where each unit of quality gain incurs an equal privacy cost, generators would follow a diagonal “naive frontier” resembling a $y = x$ line. In practice, however, models deviate from this pattern: CTGAN and GPT-4o-mini deliver moderate quality with relatively low leakage—offering better trade-offs and falling \emph{below} this idealized trend. In contrast, LLaMA 3.3 70B and especially SMOTE incur significantly higher privacy risks for their utility gains, landing \emph{above} the trend. These deviations reveal that better privacy–utility trade-offs are achievable, and that there is clear headroom for improvement.

Take-away: Foundation-model in-context synthesis can match or
exceed traditional generators in utility, but this comes at a measurable privacy cost—particularly for open-source LLMs such as LLaMA 3.3 70B—reinforcing our call for lightweight leakage-mitigation practices.



\begin{figure*}[!ht]
  \centering
  \begin{subfigure}[b]{0.32\textwidth}
    \includegraphics[width=\linewidth]{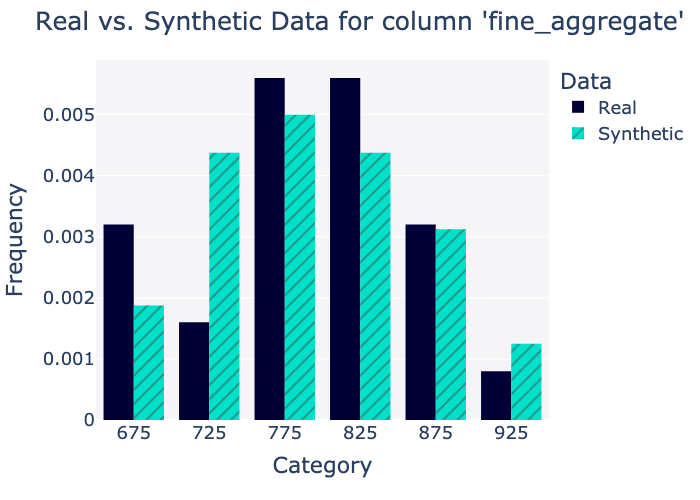}
    \caption{Default (batch = 32, temp = 1.0)}
    \label{fig:dist-default}
  \end{subfigure}
  \hfill
  \begin{subfigure}[b]{0.32\textwidth}
    \includegraphics[width=\linewidth]{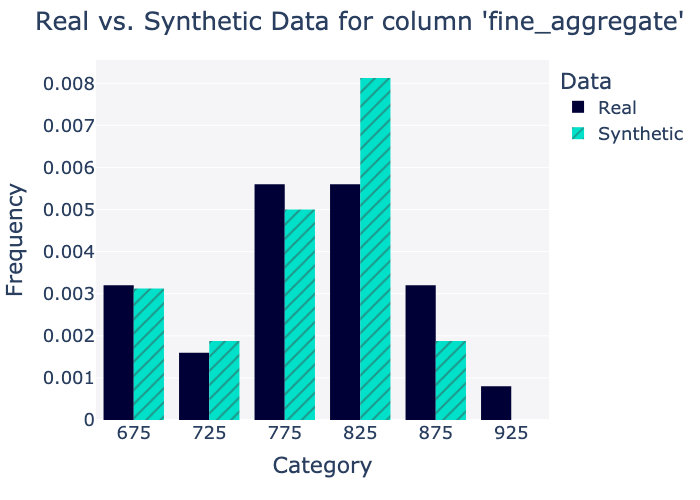}
    \caption{Reduced batch (batch = 10, temp = 1.0)}
    \label{fig:dist-batch10}
  \end{subfigure}
  \hfill
  \begin{subfigure}[b]{0.32\textwidth}
    \includegraphics[width=\linewidth]{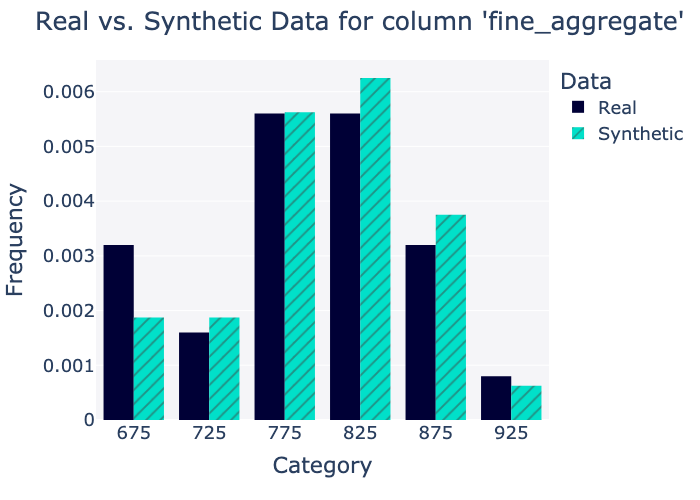}
    \caption{Low temperature (batch = 32, temp = 0.1)}
    \label{fig:dist-temp01}
  \end{subfigure}
  \hfill
  \caption{Binned value–frequency distributions for the numeric column \texttt{fine\_aggregate} from one of the 35 benchmark datasets in our CTR-23 suite, under three prompt settings for a single split (\(n=32\), seed 0), generated by LLaMA 3.3 70B.  Each panel shows the frequency of rows falling into equal-width bins of the original variable.  Comparing (a) default prompt, (b) small batch size, and (c) low temperature, highlights how these factors concentrate mass in central bins and suppresses long-tail diversity.}
  \label{fig:dist-ablation}
  
\end{figure*}

\subsection{Factorial Study of Privacy Drivers}

\begin{table*}[ht]
  \centering
  \caption{Factorial Study of Privacy Drivers Results. Each block shows the mean over four datasets with the highest average max AUC (concrete-compressive-strength, naval-propulsion-plant, solar-flare, white-wine) of the maximum attack AUC, average column-shape similarity, column correlation similarity, proportion of synthetic samples closer to real, and rare‐class ROC AUC.}
  \label{tab:ablation_summary}
  \begin{tabular}{@{} l  c  c  c  c  c @{}}
    \toprule
    \textbf{Ablation} 
      & \textbf{Max Attack AUC} 
      & \textbf{Avg.\ Shape} 
      & \textbf{Corr Similarity} 
      & \textbf{Prop.\ Closer} 
      & \textbf{Rare‐Class AUC} \\
    \midrule
    Default                          & 0.814 & 0.380 &  0.232 & 0.784 & 0.979 \\
    Batch size (k=10)                      & 0.702 & 0.340 & 0.036 & 0.727 & 0.740 \\
    Summary stats not in prompt      & 0.878 & 0.365 & 0.200 & 0.783 & 0.904 \\
    Temperature = 0.1                & 0.811 & 0.364 & 0.222 & 0.758 & 0.842\\
    Temperature = 0.5                & 0.818 & 0.392 & 0.258 & 0.779 & 0.860 \\
    \bottomrule
  \end{tabular}
\end{table*}

\vspace{0.3em}
\noindent\textbf{Findings and implications.}
Table~\ref{tab:ablation_summary} probes four single–parameter tweaks to the
\textsc{LLaMA-3.3 70B} prompting pipeline on 4 dataset with the highest average max AUC(strongest privacy leakage).

\begin{itemize}[leftmargin=1.1em,itemsep=0.3em]
\item \textbf{Smaller generation batch (10 rows).}
Reducing the batch from its 100-row default lowers the worst-case attack AUC
by 14 pp and slashes the \emph{rare-class} AUC from $0.979$ to $0.740$.  The price is an 11 pp drop in marginal-shape similarity and an almost complete collapse of correlation similarity, showing that privacy gains come mainly from a loss of diversity.

\item \textbf{Removing summary statistics from the prompt.}
Omitting global means/variances \emph{increases} leakage to $0.878$
(+0.064) while also hurting fidelity, indicating that summary statistics act
as a soft regularizer anchoring the model to population-level trends.

\item \textbf{Sampling temperature.}
Lowering the temperature to $0.1$ trims AUC only marginally
($0.814\!\rightarrow\!0.811$) but still reduces rare-class exposure
($0.979\!\rightarrow\!0.842$).
Conversely, raising it to $0.5$ nudges leakage up to $0.818$ yet yields the
best correlation similarity ($0.258$), highlighting temperature as a fine-grain
knob on the privacy–utility frontier.
\end{itemize}

\noindent\textbf{Why does attack efficacy \emph{fall} despite fewer exemplars and lower randomness?}  
At first glance, one might expect that conditioning on fewer rows (smaller batch) or sampling with a near-deterministic decoder (low temperature) would \emph{increase} lead to increased leakage: the model has less contextual diversity and may simply copy the in-context examples.  Our results show the opposite, and the key lies in how attacks exploit \emph{rare values}.  Membership-inference methods rely most heavily on rows whose quasi-identifiers are infrequent in the population; such outliers maximize the attacker’s posterior shift when observed in the synthetic set.  By contrast, high-frequency tuples offer little discriminative signal.  Both knobs we vary push the generator toward the mode of the data distribution:  

\begin{itemize}[leftmargin=1.35em,itemsep=0.2em]
\item \textbf{Smaller batch size} reduces the chance that any rare combination even appears in the prompt, biasing the model toward majority patterns seen across tasks.  
\item \textbf{Lower temperature} shrinks the softmax entropy at each decoding step, further concentrating mass on high-probability categories or central numeric quantiles.  
\end{itemize}

\noindent\textbf{Visualizing diversity collapse.}
Figure~\ref{fig:dist-ablation} presents per-category frequency histograms under three conditions: (a) the default prompt, (b) reduced batch size, and (c) low sampling temperature. Both ablations exhibit a pronounced loss of support in the distributional tails—rare bins become markedly underrepresented—while the default configuration preserves coverage across the full range of real values. This contraction of long-tail coverage closely parallels the observed decline in rare-class AUC, reinforcing the conclusion that privacy improvements arise primarily from the suppression of low-frequency events rather than from wholesale elimination of copying common patterns.

To verify this mechanism, we partition each training split into \emph{rare} and \emph{common} subsets, labelling a value as rare if its category frequency is $\le 5\%$ (categorical) or if a numeric binned into 20 equal-width intervals also falls into a $\le 5\%$ bin.  We then compute the worst-case AUC \emph{restricted to rare rows}.  As shown in the rightmost column of Table~\ref{tab:ablation_summary}, the rare-class AUC plummets from $0.979$ (default) to $0.740$ (batch\,$=10$) and $0.842$ (temperature\,$=0.1$)—drops of 24–39 percentage points that dwarf the 3–11 percentage changes in overall AUC.  The privacy gain therefore stems almost entirely from suppressing rare-value emission rather than from any reduction in verbatim copying of common rows.  This finding reframes the trade-off: \emph{privacy is improved because diversity, particularly on the long tail, is curtailed}, underscoring a fundamental tension between protecting outliers and preserving their statistical signal.

\noindent\textbf{Practical takeaway.}
Prompt-level controls—batch size, inclusion of summary statistics, and
temperature—let practitioners traverse the privacy–utility trade-off
\emph{without retraining}.
A conservative default for sensitive releases is: \textit{small batch,
low (but non-zero) temperature, and explicit summary statistics},
which cuts worst-case AUC by up to 0.11 while keeping marginal-shape
similarity within 10\% of the baseline.

\section{Conclusion}

We have presented the first systematic, end-to-end benchmark of foundation-model in-context tabular generators against data-specific GAN, VAE, and diffusion baselines under low-data regimes.  Evaluating 35 public datasets with split sizes of 32, 64, and 128 real rows, we measured statistical fidelity, downstream utility, and worst-case membership-inference leakage across 13 state-of-the-art attacks.  Our results reveal that while foundation models could achieve competitive fidelity and predictive performance, they also occupy the upper end of the privacy-risk spectrum. Simple prompt-level adjustments (batch size, summary-stat inclusion, and temperature) can shift a generator along the privacy–utility frontier by cutting leakage up at modest fidelity cost, without any retraining.

Despite these insights, our study has several limitations.  We focus solely on membership-inference as the privacy metric; other risks (attribute inference, reconstruction attacks, or linkage to auxiliary data) remain unquantified.  Second, our prompt designs and hyper-parameters cover a limited slice of the vast LLM configuration space, and future work should explore automated prompt tuning or private prompting methods. Addressing these gaps—by broadening attack models, integrating formal differential-privacy guarantees, and extending benchmarks to richer data domains—will be crucial to deploying foundation models safely in privacy-sensitive applications. As organizations increasingly turn to generative tools for data augmentation and sharing, our benchmark offers a foundational template for evaluating trade-offs between data fidelity, utility, diversity, and privacy risk in realistic, low-data contexts.

\printbibliography

\newpage

\onecolumn
\appendix


\section{Generator Implementation Details}

\noindent\textbf{TabPFN Generation~\cite{hollmann2025accurate}}:  
We follow the Prior Labs tutorial (\url{https://priorlabs.ai/tutorials/unsupervised/}). Each train split is loaded, shuffled, and batched (200 rows). Numeric features are cast to \texttt{float32}, and categoricals are label-encoded (unseen $\rightarrow$ –1). Zero-variance columns are dropped before fitting and reinserted after sampling. For each batch, we fit the unsupervised TabPFN model and sample synthetic rows with temperature $t=1.0$ across three random permutations. Outputs are decoded, constants reattached, batches concatenated, and truncated to the original row count.

\vspace{1em}
\noindent\textbf{LLaMA Generation~\cite{meta2024llama3_3_70b}}:  
Using LLaMA 3.3 70B via the Groq API (\url{https://console.groq.com/docs/models}), each train split is divided into batches of up to 32 rows. We ensure all rows are included in the prompt. Summary statistics are computed per column, and the batch is serialized to CSV. We call \texttt{llama-3.3-70b-versatile} with temperature=1.0, requesting $N$ rows as JSON. If parse errors or incorrect lengths occur, we retry up to 5 times. Valid outputs are concatenated, truncated, or re-prompted, and validated for type and dimension consistency.\footnote{LLaMA 3.3-70B failed on \textit{geographical-origin-of-music}, \textit{pumadyn32nh}, \textit{student-performance-por}, \textit{superconductivity}, and \textit{wave-energy} due to token limits; TabPFN failed on \textit{geographical-origin-of-music} due to extreme dimensionality.}

\vspace{1em}
\noindent\textbf{GPT-4o-mini~\cite{openai2024gpt4omini}}:  
Used the same prompt and inference pipeline as LLaMA 3.3 70B. We use the structured output API, with the format defined as a JSON schema where keys are column names and values are cell entries.

\vspace{1em}
\noindent\textbf{SMOTE}:  
We extend the original SMOTE algorithm to interpolate all classes. A target class is randomly sampled based on empirical frequencies, followed by interpolation using the $k=5$ nearest neighbors and $\alpha=0.5$.

\vspace{1em}
\noindent\textbf{CTGAN}:  
We use the official implementation at \url{https://github.com/sdv-dev/CTGAN} with: embedding dim = 128; generator = (256, 256); discriminator = (256, 256); learning rates = 0.0002; decay = 0.000001; batch size = 500; epochs = 300; discriminator steps = 1; pac size = 5.

\vspace{1em}
\noindent\textbf{TVAE}:  
We use default parameters from \url{https://docs.sdv.dev/sdv}: class dims = (256, 256, 256, 256); random dim = 100; 64 channels; l2scale = 1e-5; batch size = 500; epochs = 300.

\vspace{1em}
\noindent\textbf{TabDiff Generation~\cite{shi2025tabdiffmixedtypediffusionmodel}}:  
We use the default implementation with two changes: (i) early stopping if loss stagnates for 25 epochs, and (ii) relaxed preprocessing so that train splits are not required to retain every category in small datasets.

\vspace{1em}
\section{Prompt Template}
\label{sec:prompt}

\begin{lstlisting}[basicstyle=\ttfamily\footnotesize, breaklines=true, frame=single, caption={Prompt passed to Groq API}, label={lst:prompt}]
System role: You are a tabular synthetic data generation model.

Your goal is to produce data that mirrors the given examples in 
causal structure and feature/label distributions, 
while maximizing diversity.

Context: Leverage your in-context learning to generate realistic, 
diverse samples.

Output format: JSON.

Dataset name: {dataset_name}

Column names (in order): {col_names}

Summary statistics:
{summary_stats}

CSV of full data:
{data}

Please generate {batch_size} rows of synthetic data.

Treat the rightmost column as the target. Return only a JSON object:
{
  "synthetic_data": "<CSV string>"
}

Do not include any additional text.
\end{lstlisting}

\newpage

\end{document}